\documentclass[review]{elsarticle}

\usepackage{multirow}
\usepackage{amsmath}
\usepackage{upgreek}
\usepackage{amssymb}  

\journal{Journal of \LaTeX\ Templates}

\begin{document}

\begin{frontmatter}

\title{A multitask transformer to sign language translation using motion gesture primitives \tnoteref{mytitlenote}}



\author[mymainaddress]{Fredy Alejandro Mendoza López}

\author[mymainaddress]{Jefferson Rodriguez}

\author[mymainaddress]{Fabio Mart\'{\i}nez}
\ead[url]{famarcar@uis.edu.co}

\address[mymainaddress]{Biomedical Imaging, Vision and Learning Laboratory (BIVL$^2$ab). Universidad Industrial de Santander, Bucaramanga (UIS), Colombia}

\begin{abstract}

The absence of effective communication the deaf population represents the main social gap in this community. Furthermore, the sign language, main deaf communication tool, is unlettered, \textit{i.e.,} there is no formal written representation. In consequence, main challenge today is the automatic translation among spatiotemporal sign representation and natural text language. Recent approaches are based on encoder-decoder architectures, where the most relevant strategies integrate attention modules to enhance non-linear correspondences, besides, many of these approximations require complex training and architectural schemes to achieve reasonable predictions, because of the absence of intermediate text projections. However, they are still limited by the redundant background information of the video sequences. This work introduces a multitask transformer architecture that includes a gloss learning representation to achieve a more suitable translation. The proposed approach also includes a dense motion representation that enhances gestures and includes kinematic information, a key component in sign language. From this representation it is possible to avoid background information and exploit the geometry of the signs, in addition, it includes spatiotemporal representations that facilitate the alignment between gestures and glosses as an intermediate textual representation. The proposed approach outperforms the state-of-the-art evaluated on the CoL-SLTD dataset, achieving a BLEU-4 of $72,64\%$ in split 1, and a BLEU-4 of $14,64\%$ in split 2. Additionally, the strategy was validated on the RWTH-PHOENIX-Weather 2014 T dataset, achieving a competitive BLEU-4 of $11,58\%$.

\end{abstract}


\end{frontmatter}


\section{Introduction}

Approximately $1.5$ billion people have some associated degree of hearing loss worldwide. Sign language (SL) facilitates communication with deaf and hard-of-hearing people. These languages are composed of visio-spatial gestural movements and expressions, together with complex manual and non-manual interactions. Today there are more than 150 official SLs with multiple variations in each country. Like any language, there is an intrinsic grammatical richness with multiple gestural and expressive variations. These aspects make the modeling of SLs a very challenging task, even for the most advanced computer vision and representation learning methodologies. In fact, signs do not have a direct written representation, which makes it more difficult to structure the language, implying major challenges to find correspondence with other textual languages. Hence, the gloss is the principal effort to achieve a written projection, which corresponds to the word that best represents the information to be conveyed in a single sign.

Nowadays, the current state-of-the-art approaches have advanced in automatic SL translation based on encoder-decoder architectures, that include recurrent levels to model video to text correspondences \cite{camgoz2018neural}. These approaches have reported remarked results on samples with multiple variations on relative large datasets. However, these deep representations are limited to capturing mainly small and large sign dependencies that are key in SL communications. More recently, transformer approaches, have approximated the translation task by computing attentional matrices that recover complex and long-temporal interactions among signs \cite{camgoz2020sign, yin2020better}. These architectures are commonly optimized on 2D convolutional representations of raw videos by downplaying the temporal processing of the intrinsic motion \cite{rodriguez2020understanding}. However, the complex video-to-text correspondences remains challenging because multiple path posibilities. The glosses, as native written projections might be key as intermediate representation to achieve coherent translations.

This work introduces a multitask transformer architecture that include a gloss intermediate representation to achieve a more suitable translation. The proposed approach  also include a dense motion representation that enhance gestures from background and includes kinematic information, a key component in SL. These kinematic primitives are learned from a dense apparent motion fields of the gestural component of the SL sentences. These dense representations directly capture the entire gestural dynamic but also preserve the shape of the signs during the sequences. The transformer then receives video motion embedding descriptors to perform a joint end-to-end optimization, initially associating the extracted kinematic descriptors with the gloss representations in an intermediate step to generate a more aligned and accurate final textual translation.

\section{Current Work}

Multiple computational initiatives have been dedicated to support sign interpretation and translation into text sequences. Seminal approaches have proposed descriptors to capture simple image primitives and  model isolated gestures, but reporting restrictions associated to the inherent challenges such as  textural variability and also to the poor dynamic modelling of gesture changes during a translation \cite{pigou2014sign}. To overcome temporal modelling, some statistical approaches have proposed to approximate continuous SL recognition (CSLR) by tracking gestures from dominant hand \cite{koller2015continuous}. These models have generally designed under a hidden Markov model (HMM) hypothesis, managing to include some grammatical considerations and local relations in the sentences \cite{cooper2012sign}. However, these models have limitations of flexibility to include the multiple variants of language and their temporal character is generally limited.

Currently, machine SL translation are based on deep learning architectures, integrating complex encoding and decoding strategies for visual and textual information. For instance  Cui et al. \cite{cui2017recurrent} proposed a recurrent and convolutional neural network to integrate visual information (convolutional representation), and exploiting temporal information (recurrent strategy). This hybrid approach provides an approximation to capture temporal dependencies but only in short intervals, with insufficient gloss sequence matching in some language samples. Hence, Camgoz et al. \cite{camgoz2018neural} introduced the SL translation (SLT) problem, contextualized from a encoder-decoder architecture, that include recurrent, convolutional and attention modules, that together recover nonlinear dependencies during translation. This work shows strong advantages to include relative long language dependencies but their representation remain based on raw frame inputs, which makes the training and optimization process significantly costly.

Other approaches have tackle such limitation by including input representation based on movement of gestures, such as optical flow and postural inputs \cite{rodriguez2021important, de2020sign}. These motion input representations include a very important marker of language \textit{i.e., } the kinematic information, cue information to carried out more compact and therefor reliable in training schemes.  These architectures however lost intermediate representations during the video-to-text translation, resulting in some incoherent  text translations, and with some trend to overlearn projections between both sources. Cheng et al. \cite{cheng2020fully} propose an architecture consisting only of convolutional networks that also included intermediate representation and avoid exhaustive learning from a recurrent configurations. This approach take advantage of 1D convolutional representations but assume the temporal interval of each signal as an average windows that is far from high variability of such actions during a real communication. 

More recently, the transformers have been ideal architectures to deal with long term dependencies on translation strategies. These strategies have been adapted in SLT by considering deep representation that only consider attention modules to map from videos to text sequences \cite{camgoz2020sign, saunders2020progressive, yin2020better, yin2020sign}. These approaches have demonstrated remarkable capabilities to capture non-linear dependencies, and to built embedding spaces, where video and text information converge to find associations. These approaches have outperform state-of-the-art approach but remains limited to challenging real scenarios, where the inclusion of new expressions increase exponentially the complexity of the problem, because the enriched version of signs vocabulary, together with the multiple versions to develop same ideas. Hence, grouping signs in coherent glossed may reduce computational complexity of representation and can be deployed in real scenarios. In such sense, Yin and Read \cite{yin2020better} include the glosses but make the translation in two steps, getting first the glosses and based on them get the final translation, this represents a bottleneck information between the two steps, because the final translation is conditioned by the translation at gloss level. Also,  Camgoz et al. \cite{camgoz2020sign} include glosses but the representation from appearance result challenging for achieve effective translations. Additionally, the visual features used are vector embeddings from raw images.

\section{Proposed approach: A Motion-gloss SL Transformer}

\begin{figure}[h!].
\centering
\includegraphics[width=1\textwidth]{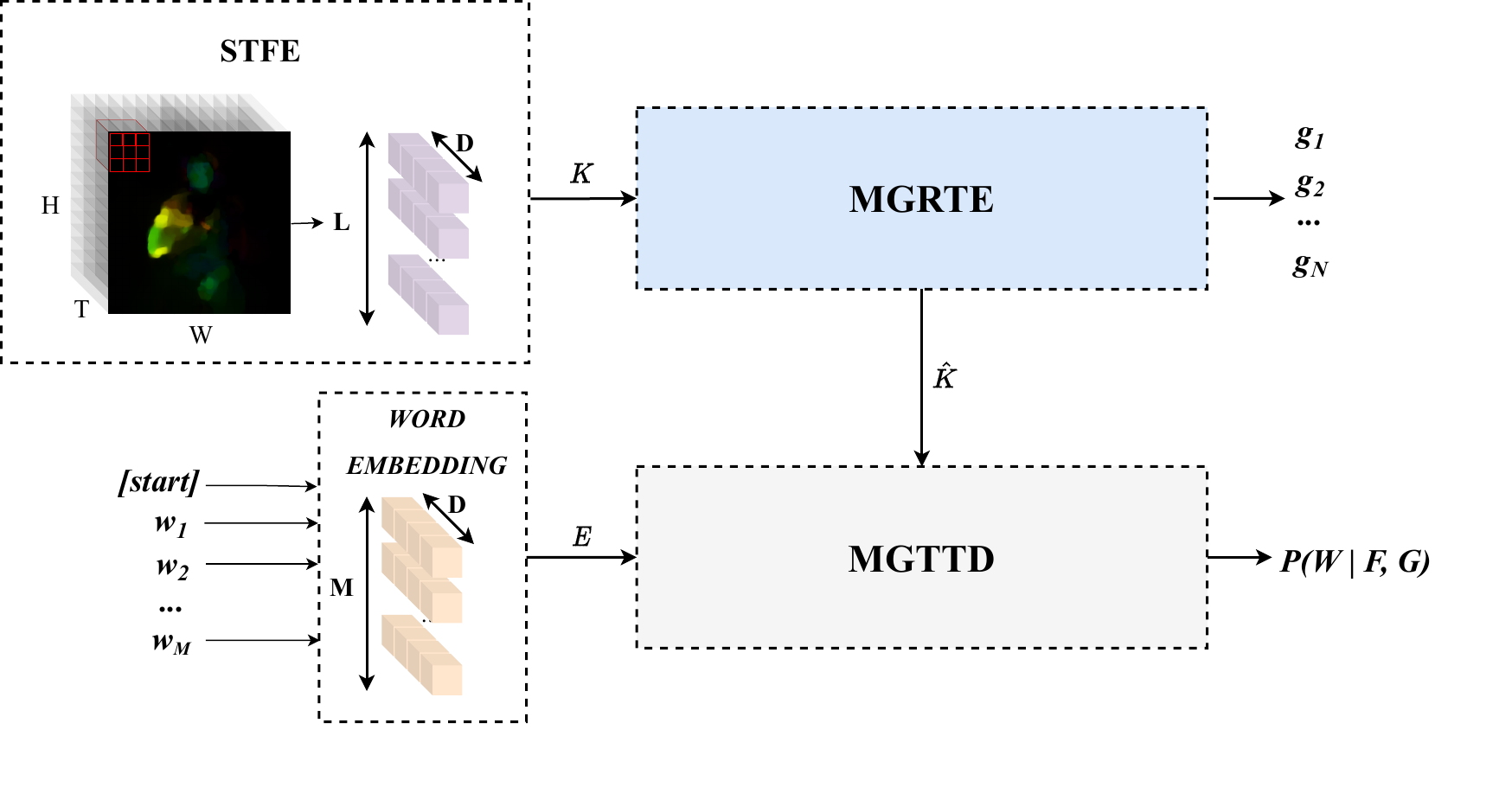}
\caption[Scheme of the proposed method.]{Proposed method. The optical flow is given as input to a volumetric strategy (STFE) that extracts at a low level the kinematic information from the video. This information is processed through the encoder (MGRTE), which, through nonlinear relationships between the spatiotemporal information, models and recognizes the glosses. On the other hand, the decoder (MGTTD) generates the translation based on temporal projections between the videos and the written language. \textbf{STFE:} Spatiotemporal Feature Extractor, \textbf{MGRTE:} Motion-gloss recognition transformer encoder, \textbf{MGTTD:} Motion-gloss translation transformer decoder.}
\label{fig:proposed_method}
\end{figure}

This work introduces a transformer that enables SL translation and gloss recognition from the optical flow of SL videos. Each video stream $\mathbf{V} = (v_1, \cdots, v_T )$ with $\mathbf{V} \in \mathbb{R}^{T \times W \times H \times C}$ represents a sequence of frames, where $T$ is the temporal length, $(W \times H)$ is the spatial resolution, and $C$ refers to the number of channels in each frame. A video corresponds to a sentence in SL that can be described with $N$ glosses as $\mathbf{G} = (g_{1}, \cdots, g_{N} )$. Likewise, each gloss sequence has a projection in a written language sentence $\mathbf{W} = (w_1, \cdots, w_M )$ with $M$ words. This indicates that expressions in SL have correspondences in three different representations: $\mathbf{V} \to \mathbf{G} \to \mathbf{W}$. The proposed strategy estimates the next word in a sentence given an intermediate gloss representation and video information; this can be defined as $P(W_t | V, G, W_{t-1})$. The general scheme of the proposed method is shown in Figure \ref{fig:proposed_method}.

\subsection{{\label{sec:kinFeat}} From video to optical flow representation}

\begin{figure}[h!]
\centering

\includegraphics[width=1\textwidth]{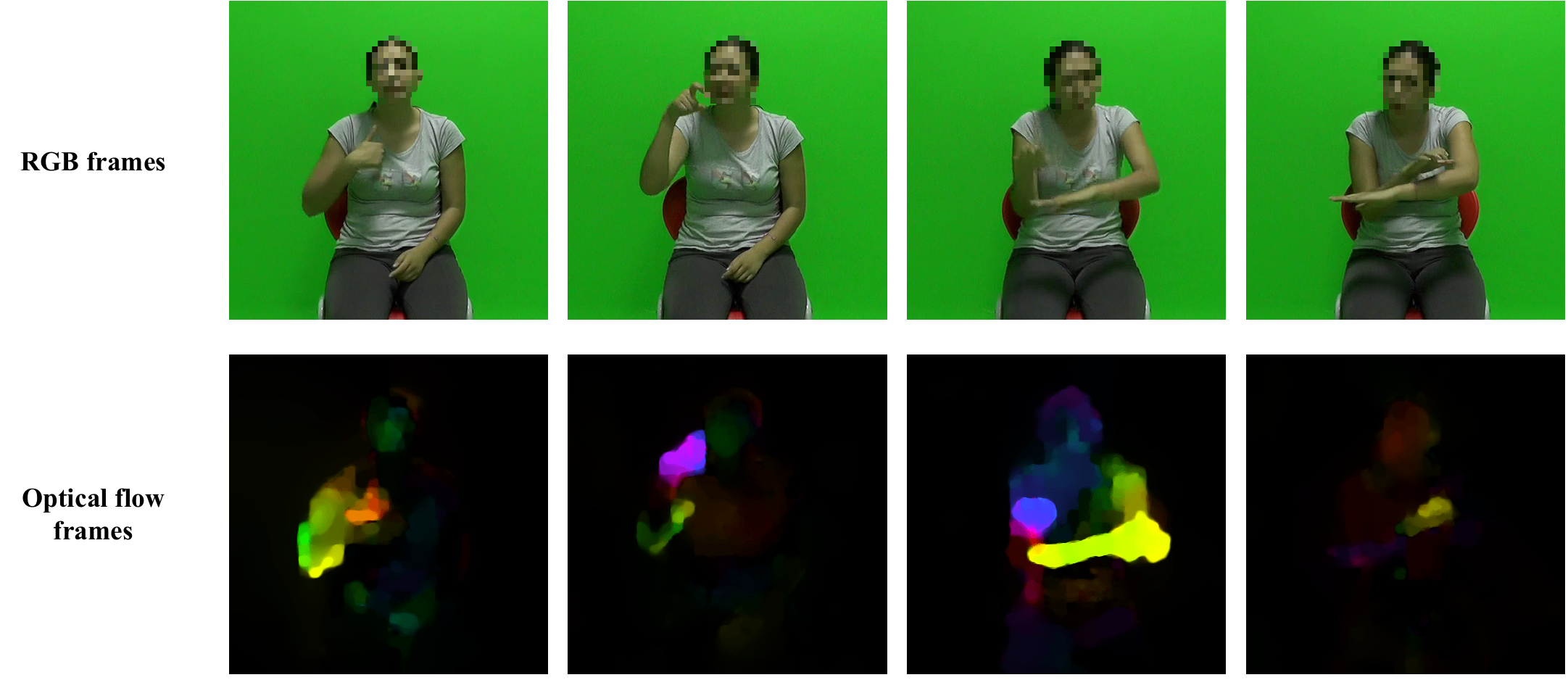}
\caption[Example of Brox optical flow representation.]{Example of Brox optical flow representation. A set of frames from an SL video represented in Brox optical flow. Each frame in this representation contains visual features that highlight motion.}
\label{fig:rgbybrox}
\end{figure}

SL is mainly based on gestural and visual components, where not only the postural geometry defines the communication, but also the movement is part of the language \cite{sandler2012phonological}. In this work we include a spatiotemporal representation that considers shape changes, and kinematic primitives for SL description. So, a video sequence $\mathbf{V} = (v_1, \cdots, v_T )$ is mapped to a kinematic representation $\mathbf{F} = (f_1, \cdots, f_T )$ to avoid the high spatial variability of SL interpreters, removing background information. Such transformation $(\mathbf{V} \to \mathbf{F})$ is achieved by computing a dense optical flow. Particularly, in this work we implemented the Brox optical flow \cite{brox2010large}, which takes into account large displacements, one of the most relevant features for SL. For this purpose, for each pair of consecutive frames $(v_{t}, v_{t+1})$ a velocity field ($nu_{t}$) resulting from a typical minimization process is computed, considering the apparent error ($E_a(\nu) = |v_{t} - v_{t+1}|^{2}$), the gradient structural error ($E_g(\nu)= |v_{t} - v_{t+1}|^{2}$) and the partial structural flow field $E_{s}(\nu) = \Uppsi|\nabla u|^{2}$. The large displacements are computed from a non-local implementation, which is based on similar patterns of velocity fields between frames from key points among consecutive frames. Then, we obtain a volumetric sequence representation with consecutive dense fields $\mathbf{F} = (\nu_1, \cdots, \nu_{T-1} ) \in \mathbb{N}^{(T-1) \times W^{'} \times H^{'}\times C^{'}}$, with $(T-1)$ dense fields, being $(W^{'}\times H^{'})$ the spatial dimension and $C^{'}$ is the number of channels. An example of a set of $\mathbf{V}$ samples in a $\mathbf{F}$ representation is shown in Figure \ref{fig:rgbybrox}.

\subsection{ {\label{sec:3DCNN}} Spatiotemporal feature extractor (STFE)}

\begin{figure}[h!]
\centering

\includegraphics[width=1\textwidth]{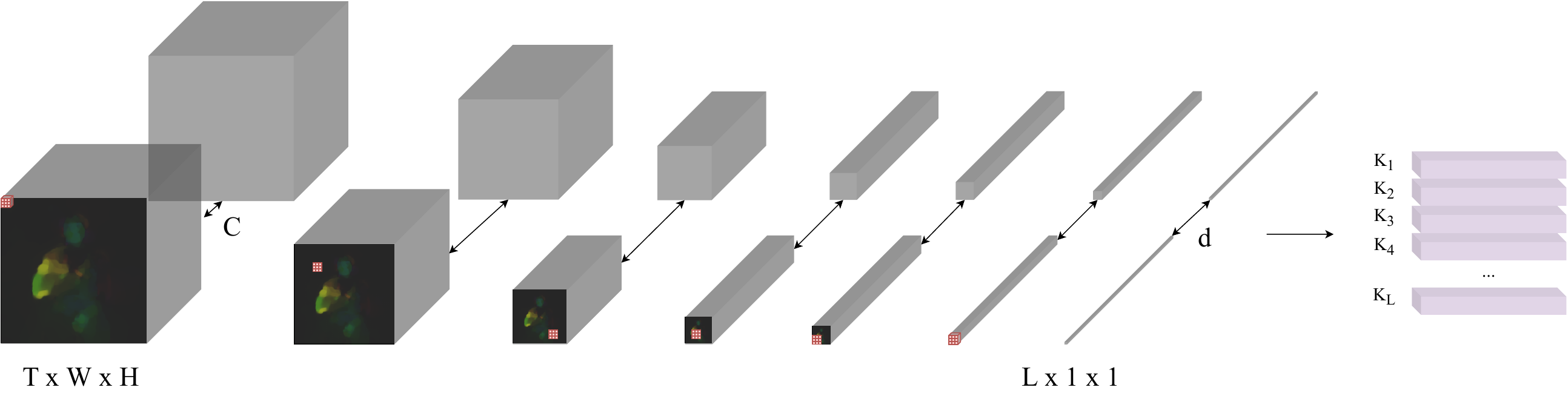}
\caption{Spatiotemporal feature extractor (STFE). This proposed module extracts low-level and long-term kinematic features using a volumetric strategy. Each sequence results in a latent space with dimensions $L \times d$ containing the most relevant information from the video.}
\label{fig:3DCNN}
\end{figure}

Furthermore a projection is made from embedding vectors that encode the gesture information $(\mathbf{F} \to \mathbf{K})$, in low dimensionality vectors $k_t$ at each time instant. The projection of the representation  $\mathbf{F} \in \mathbb{N}^{(T-1) \times W^{'}\times H^{'}\times C^{'}}$ on $\mathbf{K} \in \mathbb{N}^{L \times d}$ is achieved through a convolutional volumetric strategy, which processes volumes and progressively reduces the dimensionality down to one-dimensional descriptors, where each embedded vector $L \in \mathbb{R}^{d}$. From a 3D convolurtional net (\textit{long-term temporal convolutions} \cite{varol2017long}), the strategy learns nonlinear transformations that progressively compute a complex representation expressed in high-level projections. This strategy guarantees the quantification of kinematic information from the embedded space $\mathbf{K}$ that robustly models the spatiotemporal information of the language. Figure \ref{fig:3DCNN} shows the implemented volumetric strategy, which delivers hidden embedding vectors that quantify the complex relationships present in the kinematic representation.

\begin{figure}[h!]
\centering

\includegraphics[width=0.6\textwidth]{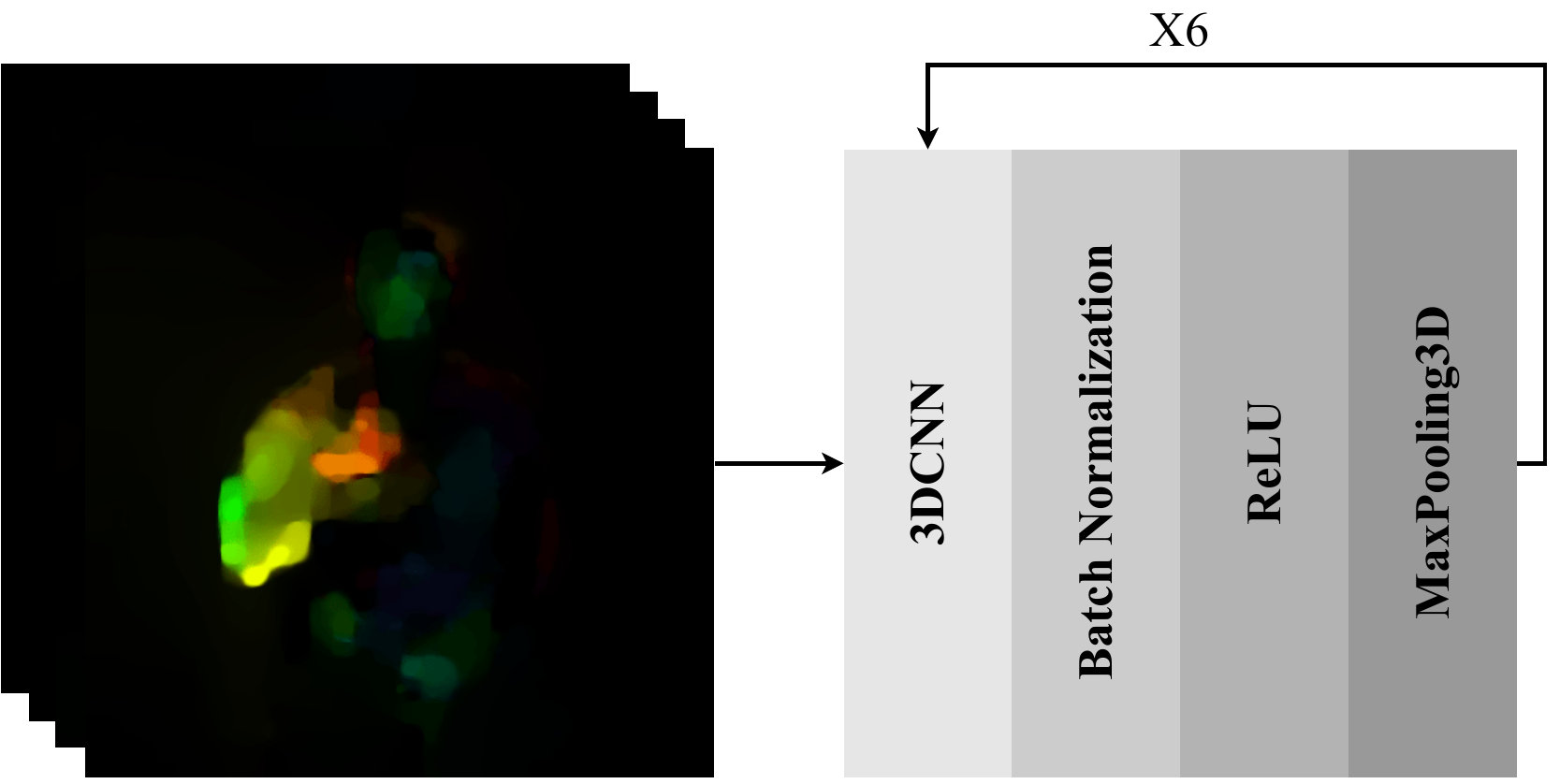}
\caption{STFE convolutional block. Each block is composed of four layers that fulfill different functions. In the proposed methodology, six consecutive blocks were used.}
\label{fig:convolutional_block}
\end{figure}

As shown in figure \ref{fig:convolutional_block}, the architecture is composed of six 3D convolutional blocks, where each block consists of a sequence of defined layers, whose order is \textit{3DCNN} $\to$ batch normalization $\to$ ReLU $\to$ max pooling 3D. For this purpose, the 3DCNN layer obtains dense representations over the $\mathbf{F}$ sequence. The batch normalization regularizes the embedded information, while the ReLU activation allows the network to determine nonlinear patterns, with bias in the positive representations. Likewise, the \textit{max pooling 3D} operation reduces the dimensionality of the tensor, while preserving the relevant information. In this case, the temporal dimension of the input videos $T$ has a different length concerning $L$.

\subsection{{\label{sec:encoder}} Motion-gloss recognition transformer encoder (MGRTE)}

\begin{figure}[h!]
\centering

\includegraphics[width=1\textwidth]{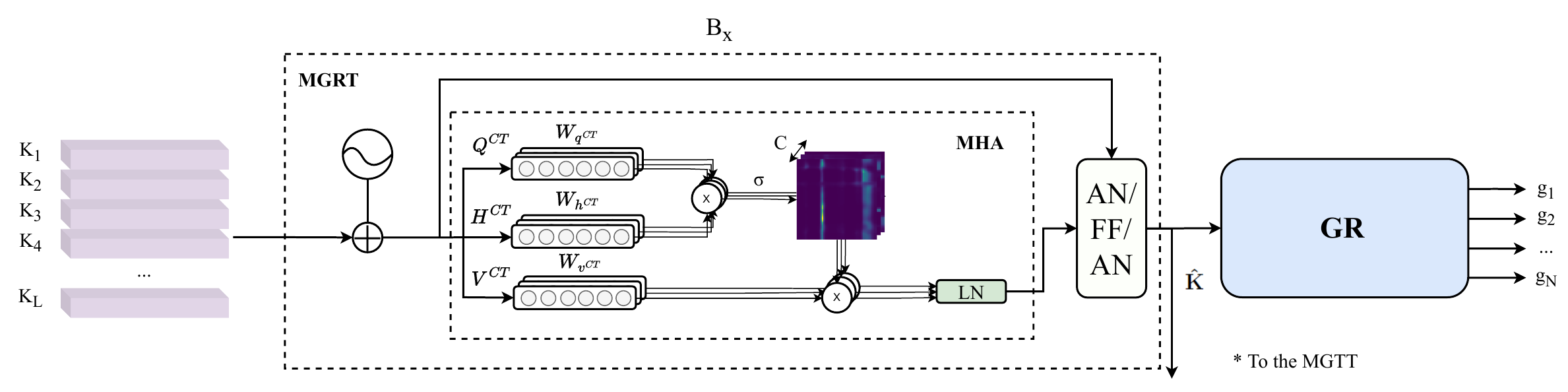}
\caption{Motion-gloss recognition transformer. This module determines a set of embedded $\mathbf{\hat{K}}$ of encoded information and a set $\mathbf{G}$ of glosses. For this, with the embedded space $\mathbf{K}$ as input, a \textit{positional encoding} layer inserts positional information to each $K_l$ vector, then, this representation passes through an attention module multi-head, which finds nonlinear relationships between each element of $\mathbf{K}$. Finally, the information is given as input to a layer \textit{point-wise feed Forward} followed by normalization modules. This sequence is repeated $B$ times, thus obtaining a matrix $\mathbf{\hat{K}}$ with encoded kinematic information. \textbf{MHA:} Attention \textit{multi-head}, \textbf{AN:} Addition and normalization, \textbf{FF:} Layer, \textit{point-wise feed-forward}, \textbf{LN:} Linear layer, $\mathbf{sigma:}$ Softmax Activation , \textbf{GR:} Gloss recognition module. }
\label{fig:CT}
\end{figure}

This module establish relationships between kinematic information embedded in a low-dimensional representation and written glosses ($\mathbf{K} \to \mathbf{G}$). For this purpose, a transformer encoder was introduced to model and recognize written glosses given an embedded kinematic representation of the input video, expressed as $P(g_1, \cdots, g_N | k_1, \cdots k_L)$. Therefore, this module acts as an encoder of $\mathbf{K}$ spatiotemporal information and intermediate representations of $\mathbf{G}$ in the SL. Figure \ref{fig:CT}, illustrates the main components used in this module, where layers with multiple self-attentions enrich the temporal representation of the embeddings and are projected through a nonlinear neural representation.

In this work, absolute positional encoding \cite{vaswani2017attention} was used for guarantee a unique value for each vector. Therefore, the process of adding temporal information to each $\mathbf{K}$ embedding is defined as: $\mathbf{K} = \mathbf{K} + PositionalEncoding(\mathbf{K})$.

\subsubsection{{\label{sec:MHAE}} Multi-head attention mechanisms (MHA) from a kinematic gloss domain}

The gloss representation from embedding vectors is achieved by projecting $\mathbf{K}$ into multiple attention mechanisms (MHA). Specifically, MHA modules determine nonlinear relationships between two domains to obtain multiple attentions that recover nonlocal relationships between the two sequences. In this case, the transformer methodology proposes to apply these attention modules on a single domain, seeking to correlate each element $K_l$ with the others of the sequence. This type of attention is called self-attention. Thus, a multiple-attention block corresponds to the parallel implementation of several self-attention mechanisms at the same processing level.

\begin{figure}[h!]
\centering
\includegraphics[width=1\textwidth]{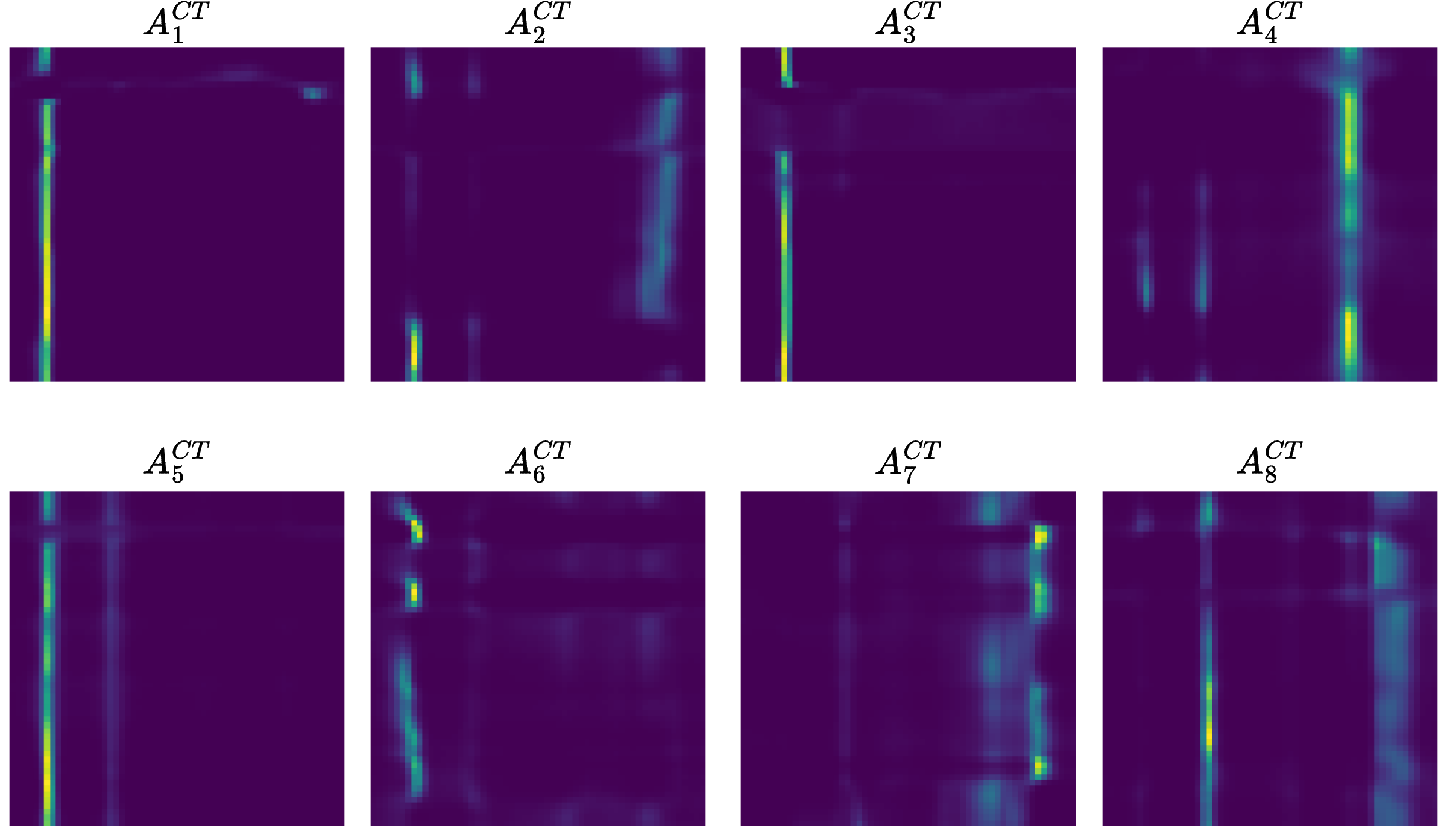}
\caption{Attention maps of $\mathbf{A}^{TE}$ glosses based on the kinematic information $\mathbf{K}$. For each head, a map with different relationships between the sequence is determined. This example has eight heads and two coding layers. The extracted information corresponds to the second layer.} 
\label{fig:MHA_encoder}
\end{figure}

Particularly, in this work, a multi-attention layer with $C$ self-attention modules was implemented. For each attention module $\mathbf{O}_i^{TE}$, the embedding vectors $\mathbf{K}$ are projected into three different components, looking for model glosses from the obtained kinematic information. These projections, typically known as: \textit{key} ($\mathbf{H}_{i}^{TE} = W_{h^{TE}} \mathbf{K}$) and \textit{query} ($\mathbf{Q}_{i}^{TE} = W_{q^{TE}} \mathbf{K}$) by projecting kinematic inputs $\mathbf{K}$ through the weights $W_{h^{TE}}$ and $W_{q^{TE}}$, respectively, where $W_{h^{TE}}, W_{q^{TE}}, \in \mathbb{R}^{L \times \frac{d}{C}}$. With these projections it is possible to compute attention maps $\mathbf{A}_i^{TE}= \sigma((\mathbf{H}_{i}^{TE})^{T} \mathbf{Q}_{i}^{TE})$ encoding nonlinear correlations between the embedded vectors, where $\sigma$ is the activation function softmax. That is, each of the attention maps $\mathbf{A}_i^{TE}$ has a nonlinear representation that defines the main relationships between two sequences, where temporal dependencies are rescued regardless of the temporal distance between elements. Figure \ref{fig:MHA_encoder} illustrates an example of the attention matrices $\mathbf{A}_i^{TE}$ obtained in this module.

On the other hand, in parallel with the projections $(\mathbf{H}^{TE}, \mathbf{Q}^{RT})$  we made a projection \textit{value:} ($\mathbf{V}_i^{RT}=W_{v^{TE}} \mathbf{K}; W_{v^{TE}} \in \mathbb{R}^{L \times \frac{d}{C}}$) which is weighted with respect to the attention matrix $\mathbf{A}_i^{TE}$. With this weighting, a new representation of embedded vectors is obtained, resulting in a feature map $\mathbf{O}_i^{TE}$ that highlights the main local and nonlinear components with the highest correspondence to the glosses. This representation is computed as $\mathbf{O}_{i}^{TE} = \mathbf{A}_{i}^{TE}\mathbf{V}_{i}^{TE}$;  $\mathbf{O}_{i}^{TE} \in \mathbb{N}^{L \times \frac{d}{C}}$. For a block of $C$ attentional mechanisms, we perform this process on multiple independently trained mechanisms, thereby recovering $C$ kinematic representations of attention.$\{\mathbf{O}_{1}^{TE}, \mathbf{O}_{2}^{TE} \ldots, \mathbf{O}_{C}^{TE} \}$. During training, multiple representations enrich the representation of the sequences, making each attentional mechanism focus on specific nonlinear relationships that contribute to the proper correspondence between kinematic vectors from video sequences and glosses. These kinematic representations are concatenated and projected to a representation vector ($MH^{TE} = \text{concat}(\mathbf{O}_{1}^{TE}, \mathbf{O}_{2}^{TE} \ldots, \mathbf{O}_{C}^{TE} )W_{Mh^{TE}}$), according to a weight setting $W_{Mh^{TE}}$. These representations are given as input through two linear dense layers, obtaining a final representation $\mathbf{\hat{K}} \in \mathbb{R}^{L \times d}$, where $\mathbf{\hat{K}}$ represents the enhanced embedded vectors through the projection of the MHA module. This processing represents a  $B_{x}^{TE}$ layer of transformer type encoding in this work. This layer can be sequentially replicated for further processing and data representation.

\subsection{{\label{sec:RGModulo}} Gloss recognition module (GR)}

This work obtains projections between the $\mathbf{K}$ and $\mathbf{G}$ domains, looking for get a better and simple alignment between the videos and the written glosses. In this sense, a written gloss unit $g_n = \{V_i\}_{i:i+T'} \sim \{\hat{K_i}\}_{i:i+L'}$ describes a text as a subset of $\Delta T$ frames, which can be approximated by a subset of $\Delta L$ in our representation $\mathbf{\hat{K}}$.

\begin{figure}[h!]
\centering
\includegraphics[width=1\textwidth]{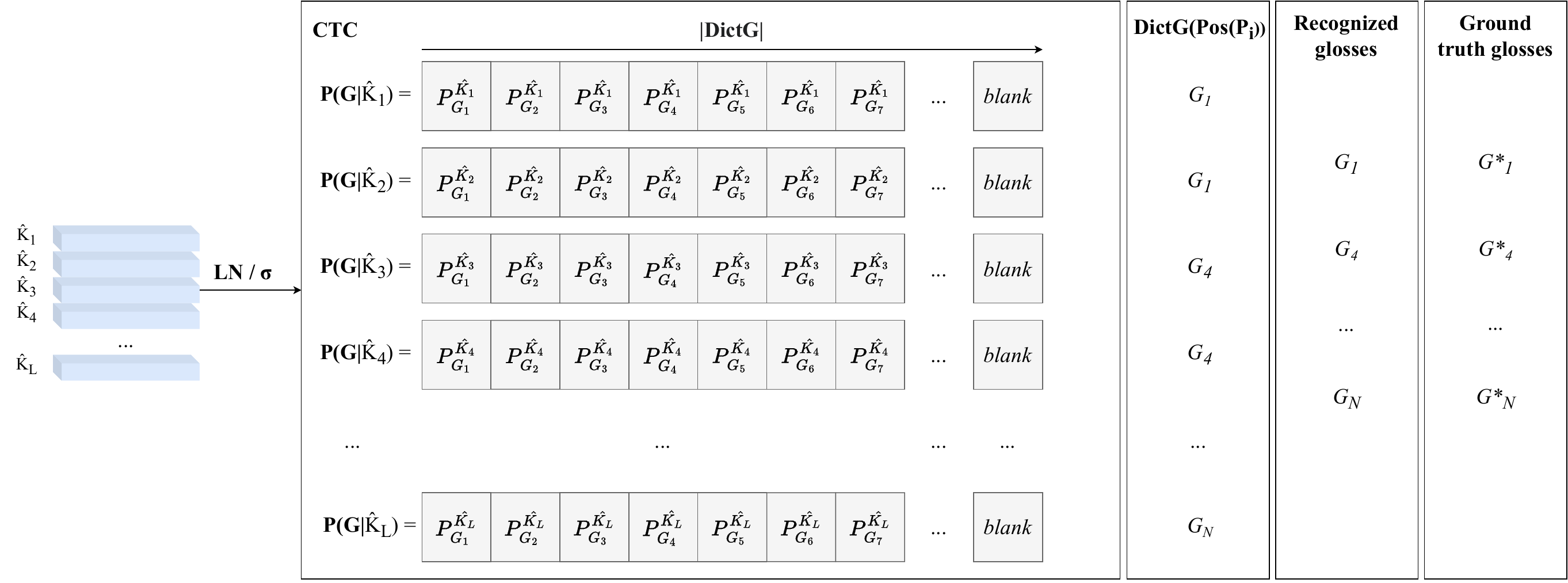}
\caption{Connectionist Temporal Classification (CTC) Module. The embedding vectors $\mathbf{\hat{K}}$ passes through a linear layer (LN) and a softmax activation ($\sigma$). The sequence of highlighted boxes corresponds to a path $\pi \in \beta$; in some steps, the module does not recognize any gloss, therefore, the special token $blank$ used. Then, from the gloss dictionary and the index of the selected dimension, a written gloss is obtained for each time step $l$. Finally, a decoding process is implemented to delete repeated glosses and characters marked as $blank$.}
\label{fig:CTC_training}
\end{figure}

To achieve this correspondence ($\mathbf{\hat{K}} \to \mathbf{G}$), in this work we use the CTC loss \cite{graves2006connectionist}, which allows to determine glosses without annotations in the temporal dimension. The objective of this module is to recognize a set of ground truth glosses $\mathbf{G^*}$, given the embedded representation $\mathbf{\hat{K}}$, defined as:  $P(\mathbf{G^{*}}| \mathbf{\hat{K}})$. In this case, this projection was implemented as the most probable path between $\hat{K}_i$ vectors and the gloss dictionary, as: $P(\mathbf{G^{*}}| \mathbf{\hat{K}}) = \sum_{\pi \in \beta}P(\pi|\mathbf{\hat{K}})$, where $\pi$ represents a path belonging to the set of $\beta$ paths that can be used to get the target glosses $\mathbf{G^{*}}$. Figure \ref{fig:CTC_training} shows how the CTC operates on a set $\mathbf{\hat{K}}$ of embedding vectors. In this case $\mathbf{\hat{K}}$ is projected onto a linear layer + softmax, which is intended to adjust the dimensionality according to the size of the gloss dictionary and to establish a gloss-level probability. Then, for each embedding, the corresponding probability is obtained concerning each gloss defined in the dictionary (rows  $P(G|\hat{K}_i)$). The most probable path  $\pi$ then corresponds to the glosses with the highest probability in each embedding. The error of the MGRTE module is determined from: $\mathbb{L}_{TE} = 1 - P(\mathbf{G^{*}}| \mathbf{\hat{K}})$. On the other hand, it is important to note that the $\mathbf{\hat{K}}$ is a encoded representation of glosses with kinematic information, which allows for relate in an MHA module the written domain $\mathbf{W}$ to $\mathbf{F}$.

\subsection{{\label{sec:decoder}} Motion-gloss translation transformer decoder (MGTTD)}

\begin{figure}[h!]
\centering

\includegraphics[width=1\textwidth]{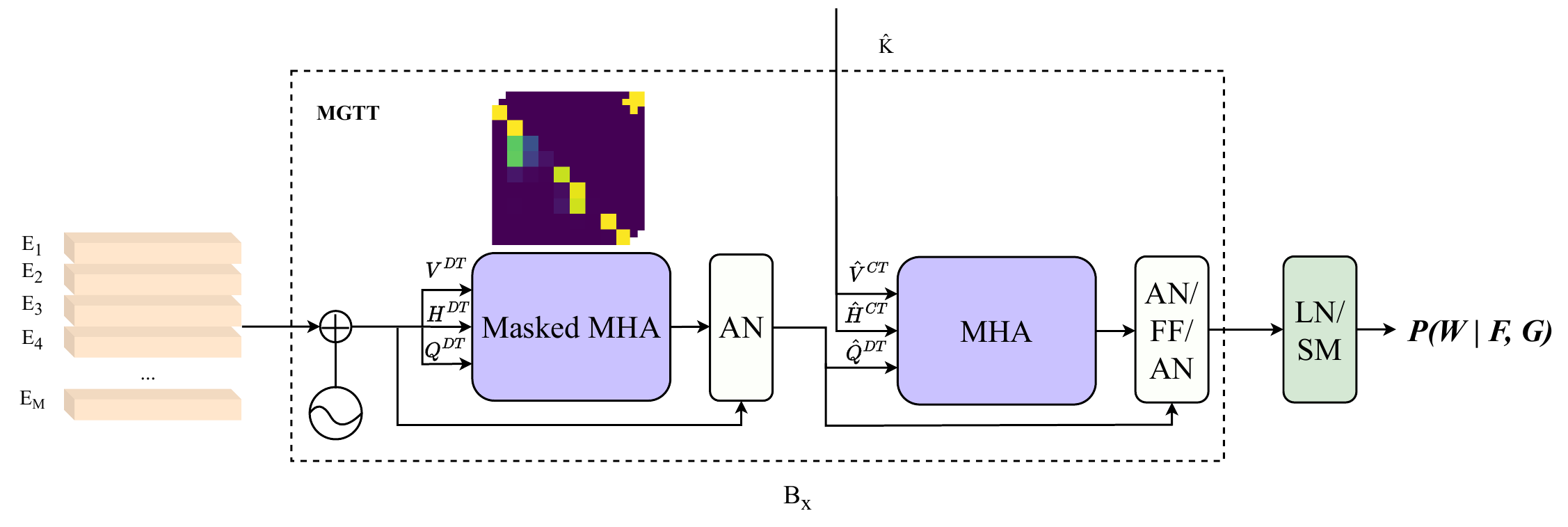}
\caption{Motion-gloss translation transformer decoder. This module aims at predicting a set of words $\mathbf{W} = (w_2 \cdots \ w_M)$. For this purpose, a latent space $\mathbf{E}$, which is an embedded representation of the words $\mathbf{W}$, is given as input to the MGTTD module, which exploits temporal relationships between the elements of $\mathbf{W}$ and determines projections between $\mathbf{F}$ and $\mathbf{W}$ through MHA modules and the information encoded in $\mathbf{\hat{K}}$. This sequence is repeated $B$ times, where finally the computed latent space passes through a linear layer and a softmax layer, outputting $P(w_m | F, G, w_{m-1})$ at each decoding step.}
\label{fig:DT}
\end{figure}

The proposed encoder allows to project and obtains an embedded representations $\mathbf{\hat{K}}$ and written glosses $\mathbf{G}$, given an input sequence $\mathbf{V} \to (\mathbf{\hat{K}}, \mathbf{G} )$. In this work, a decoder module transformer was implemented, which seeks to model the word domain  $\mathbf{W}$ and find projections between the video and the written language using the intermediate representations $(\mathbf{\hat{K}}, \mathbf{G} )$. The main modules that compose the transformer decoder implemented in this work are detailed below (an illustration is available in Figure \ref{fig:DT}).

\subsubsection{{\label{sec:WE}} Word embedding layer}

\begin{figure}[h!]
\centering

\includegraphics[width=1\textwidth]{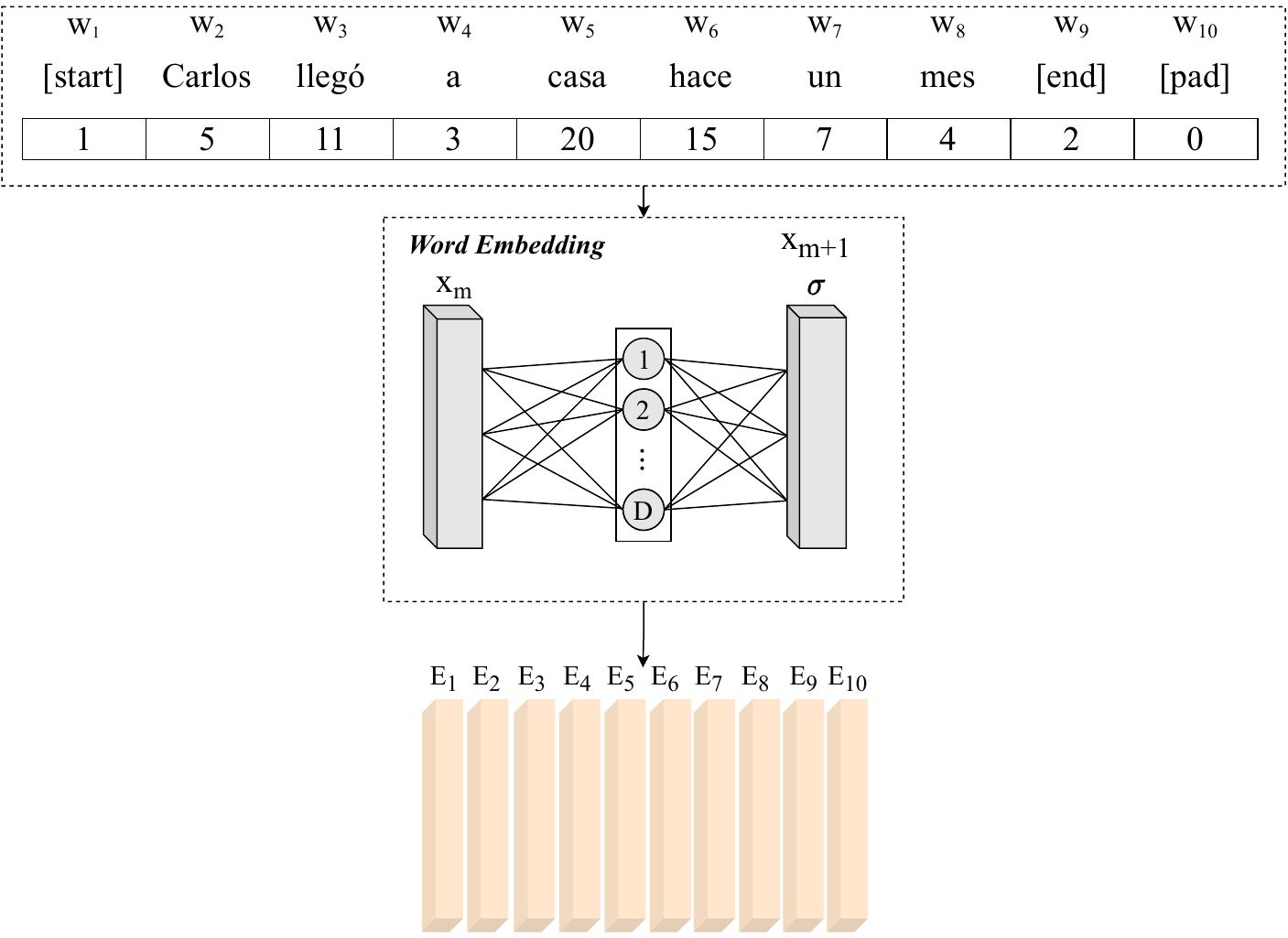}
\caption{Example of word embedding on a $\mathbf{W}$ sentence. Each word of the written sentence is assigned to a numerical token. At the $m$ decoding step the first $m$ words are concatenated, represented as $x_{m}$, which are given as an input to a linear layer with dimension $D$ that aims to find relations concerning the next word. In this case $D = 128$ and $M = 10$. The output is a set $\textbf{E}$ of embedded vectors with dimension $M \times D$.}  
\label{fig:embedding_layer}
\end{figure}

In this module, a latent space representation is obtained from $\mathbf{W}$. This latent space allows obtaining word descriptors that encode semantic information from embedded projections, \textit{i.e.}, words are positioned according to their meaning. In this case, the special words (tokens) $[start]$ and $[end]$ are added to each sentence $\mathbf{W}$ at the beginning and end of each. Also, a \textit{padding} process is performed on the sentences $\mathbf{W}$, seeking that all written sequences have the same length, as well as their respective numerical representation for each element ${w_{m}}$. Then, the embedded representation $\mathbf{E} = WordEmbedding(W)$ is implemented as a simple dense neural network, with a single intermediate layer (support of the embedded vectors) and an output layer that learns the probability of occurrence of neighboring words. In this case, the projection of each sentence, resulting in an embedded encoding $\mathbf{E} \in \mathbb{R}^{M \times D}$ which corresponds to the representation of $\mathbf{W}$, as illustrated in Figure \ref{fig:embedding_layer}. 

Positional information is added to each element of $\mathbf{E}$, using the positional encoding strategy, as: $\mathbf{E} = \mathbf{E} + PositionalEncoding(\mathbf{E})$.

\subsubsection{{\label{sec:MHACross}} MHA between SL and write language.}

The written language model is determined from MHA projections. This modules determines a $\mathbf{O}_i^{TD}$ features that recover nonlinear temporal relationships between the elements of the $\mathbf{E}$. For this purpose, the projections \textit{key}  ($\mathbf{H}_{i}^{TD} = W_{h^{TD}} \mathbf{E}$), \textit{query} ($\mathbf{Q}_{i}^{TD} = W_{q^{TD}} \mathbf{E}$) and \textit{value} ($\mathbf{V}_{i}^{TD} = W_{v^{TD}} \mathbf{E}$) are given as input to a module of multiple self-attentions, where the weights $W_{q^{TD}}$, $W_{h^{TD}}$, $W_{v^{TD}} \in \mathbb{R}^{M \times \frac{d}{C}}$. Likewise, the attention maps $\mathbf{A}_i^{TD}= \sigma((\mathbf{H}_{i}^{TD})^{T} \mathbf{Q}_{i}^{TD})$ quantify from different perspectives the temporal relationships that exist between the words. Subsequently, we determine the features $\mathbf{O}_{i}^{TD} = \mathbf{A}_{i}^{TD}\mathbf{V}_{i}^{TD}$; $\mathbf{O}_{i}^{TD} \in \mathbb{N}^{M \times \frac{d}{C}}$, and $MH^{TD} = \text{concat}(\mathbf{O}_{1}^{TD}, \mathbf{O}_{2}^{TD} \ldots, \mathbf{O}_{C}^{TD} )W_{Mh^{TD}}$, according to a weight adjustment $W_{Mh^{TD}}$ and a single representation $\mathbf{\hat{E}}$ is obtained through two linear layers. 

\begin{figure}[h!]
\centering

\includegraphics[width=1\textwidth]{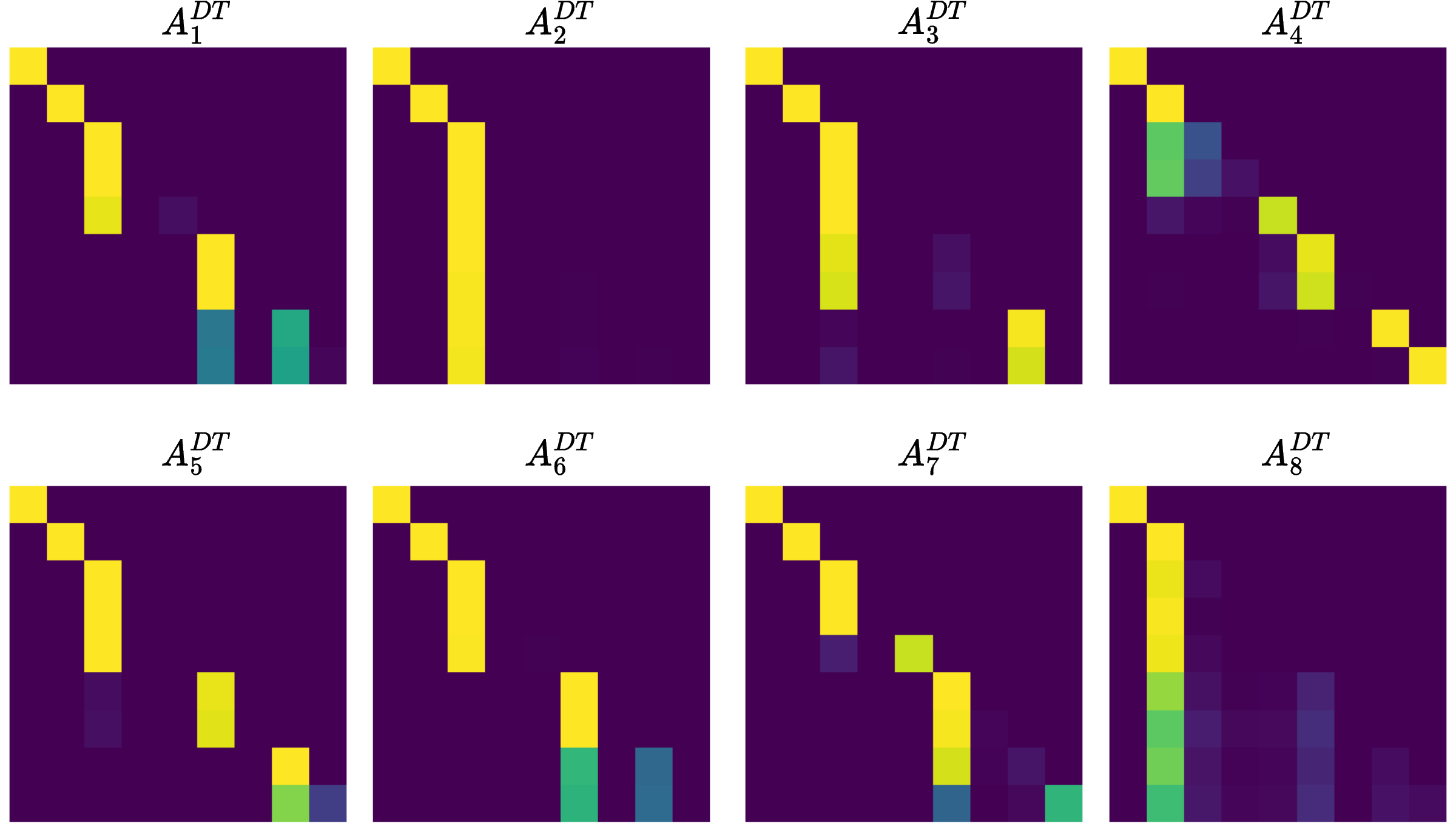}
\caption{Written language attention maps $\mathbf{A}^{TD}$. Note how the elements outside the lower triangular matrix do not correlate with any other. This example shows the attention maps of the eight heads from the second layer.} 
\label{fig:MHA_decoder}
\end{figure}

A variation in the MHA module is employed, due to that it is necessary to use a mask over the attention matrices, because at decoding step $m$, the module can only determine relations between words decoded up to step $m-1$, thus guaranteeing correct modeling of the language. In this way, the limitations in the training stage, in terms of relations between already decoded words, can be contained. For this purpose, the mask only takes into account the elements of the lower triangular matrix, weighting with $0$ the other elements of the set. Figure \ref{fig:MHA_decoder} shows an example with the attention matrices obtained in this module.

The information in $\mathbf{\hat{E}}$ contains temporal relations of the sequence of words $\mathbf{W}$, however, to perform a translation at the written language level it is necessary to find projections between the video and the text $(\mathbf{F} \to \mathbf{W})$. In this sense, a second MHA module finds projections between the two domains by exploiting temporal information between the words and the information encoded in $\mathbf{\hat{K}}$ (encoder output). This operation represents an important process, because is the step where representational relations are learned using the gloss information. For this purpose, the representations $\mathbf{\hat{H}}_{i}^{TE} = W_{\hat{h}^{TE}} \mathbf{\hat{K}}$, $\mathbf{\hat{V}}_{i}^{TE} = W_{\hat{v}^{TE}} \mathbf{\hat{K}}$ and $\mathbf{\hat{Q}}_{i}^{TD} = W_{\hat{q}^{TD}} \mathbf{\hat{E}}$, determine a  features$\mathbf{\hat{O}}^{TD}$, where from the attention maps $\mathbf{\hat{A}}_i^{TD}= \sigma((\mathbf{\hat{H}}_{i}^{TE})^{T} \mathbf{\hat{Q}}_{i}^{TD})$ and the concatenation of the features $\mathbf{\hat{O}}_{i}^{TD} = \mathbf{\hat{A}}_{i}^{TD}\mathbf{\hat{V}}_{i}^{RT}$;   $\mathbf{\hat{O}}_{i}^{TD} \in \mathbb{N}^{M \times \frac{d}{C}}$ obtain a single representation$\mathbf{\Tilde{E}} \in \mathbb{R}^{M \times D}$. 

The $\mathbf{\Tilde{E}}$ embedding vectors are given as input to a linear layer with softmax activation, where the tensor shape is adjusted according to the written words dictionary size and a probability distribution is obtained at word level, obtaining $P(w_m | F, G, w_{1:m-1})$. In addition, the decoding includes two multiple attention modules, which constitute a single processing layer$B_{x}^{TD}$.

As for the decoder training, the loss function cross entropy was used. For this, we first calculate the probability of the words given the video and glosses, which is defined as 
\[
p(W|\Tilde{E}) = \prod_{m=1}^{M} p(w_m | \Tilde{e}_m).
\]
The loss is obtained as:
\[
\mathbb{L}_{TD} = 1 - \prod_{m=1}^{M} \sum_{j=1}^{J} P(w_m^{*j}) P(w_m^{*j}|\Tilde{e}_m),
\]
where $J$ represents the size of the written language word dictionary, and $P(w_m^{*j})$ represents the probabilities of the target word $w^{*j}$ at the $m$-th decoding step.

Finally, in order to obtain gloss recognition and SL translation in a single strategy, joint training was used in the proposed method. The loss functions of the MGRTE ($\mathbb{L}_{TE}$) and MGTTD ($\mathbb{L}_{TD}$) modules are used, where each is weighted by a constant $\lambda$, which quantifies the relevance of each loss in the training stage. Therefore, the overall loss function is defined as: $\mathbb{L} = \lambda_{TE}\mathbb{L}_{TE} +  = \lambda_{TD}\mathbb{L}_{TD}$.

\section{Results}

\subsection{Experimental setup.}

A video resolution and frames were defined, which corresponds to (128 $\times$ 227 $\times$ 227 $\times$ 227 $\times$ 3). The table \ref{tab:tensor-method} specifies how the tensor shape is affected in each of the stages of the proposed strategy. A symmetric architecture (same number of layers $B_x$ for the encoder and decoder) was taken into account and experiments were performed with 1, 2, 3 and 4 layers.

\begin{table}[h!]
\centering
\begin{tabular}{|l|c|}
\hline
\multicolumn{1}{|c|}{\textbf{Module}}  & \textbf{Output shape} \\ \hline
Input layer of $\mathbf{F}$                 & (120 $\times$ 227 $\times$ 227 $\times$ 3)           \\ \hline
STFE                         & (58 $\times$ 128)            \\ \hline
MGRT                          & (58 $\times$ 128), (58 $\times$ 91)            \\ \hline
Input layer of $\mathbf{W}$                       & (12)          \\ \hline
\textit{Word embedding}          & (12 $\times$ 128)              \\ \hline
MGTT                                & (12 $\times$ 115)          \\ \hline
\end{tabular}
\caption{Tensor shape along the proposed method. The initial dimension of the video corresponds to $(T \times W \times H \times C)$.} 
\label{tab:tensor-method}
\end{table}

Table \ref{tab:parameters-SFTE} summarizes the proposed architecture and the hyperparameters used in the STFE module. This strategy allows get complex descriptors, which summarize the information observed in the $2d+t$ sequences. In this architecture, we start with three-channel video sequences, where this input is projected through layers implementing local 3D convolutions, which increase the channel depth while reducing the spatial dimensionality. Finally, the architecture allows to obtain embedding vectors that represent the video.

\begin{table}[h!]
\centering

\begin{tabular}{|c|c|c|c|}
\hline
\textbf{Layer}      & \textbf{Stride size} & \textbf{Kernel size} & \textbf{Output shape} \\ \hline
Input   & -                         & -                         & (128, 227, 227, 3)           \\ \hline
Block 1 & (2, 2, 2)                 & (3, 3, 3)                 & (63, 57, 57, 64)             \\ \hline
Block 2 & (3, 3, 3)                 & (1, 1, 1)                 & (62, 28, 28, 32)             \\ \hline
Block 3 & (3, 3, 3)                 & (1, 1, 1)                 & (61, 14, 14, 64)             \\ \hline
Block 4 & (3, 3, 3)                 & (1, 1, 1)                 & (60, 7, 7, 64)               \\ \hline
Block 5 & (3, 3, 3)                 & (1, 1, 1)                 & (59, 3, 3, 128)              \\ \hline
Block 6 & (3, 3, 3)                 & (1, 1, 1)                 & (58, 1, 1, 128)              \\ \hline
Reshape & -                 & -                 & (58, 128)              \\\hline
\end{tabular}
\caption{Tensor shape through the STFE module. The parameters for the max pooling 3D correspond to a $(3 \times 3 \times 3)$ pool size and a stride size of $(2 times 2 times 2 times 2)$.}
\label{tab:parameters-SFTE}
\end{table}

On the other hand, table \ref{tab:hyperparameters} shows a summary of the hyperparameters used for the training stage.

\begin{table}[h!]
\centering

\begin{tabular}{|l|c|}
\hline
\multicolumn{1}{|c|}{\textbf{Hyperparameter}}  & \textbf{Valor} \\ \hline
Batch size                              & 1           \\ \hline
Neurons in feed forward layer & 2048           \\ \hline
MHA Dropout                       & 0,1            \\ \hline
Epochs            & 30             \\ \hline
Optimizer                                  & Adam           \\ \hline
Learning rate                        & 1 $\times 10^{-4}$  \\ \hline
\end{tabular}
\caption{Hyperparameters used in the proposed method. After the fifth epoch, the learning rate decreases according to: $lr = lr*e^{-0.1}$.} 
\label{tab:hyperparameters}
\end{table}

\subsection{Datasets}

The proposed method was extensively evaluated on two different datasets dedicated to SL translation and gloss recognition. The two datasets are public and are described below:

\subsubsection{CoL-SLTD: Colombian Sign Language Dataset.}\label{subsec-CoL-SLTD}

\begin{figure}[h!]
\centering

\includegraphics[width=1\textwidth]{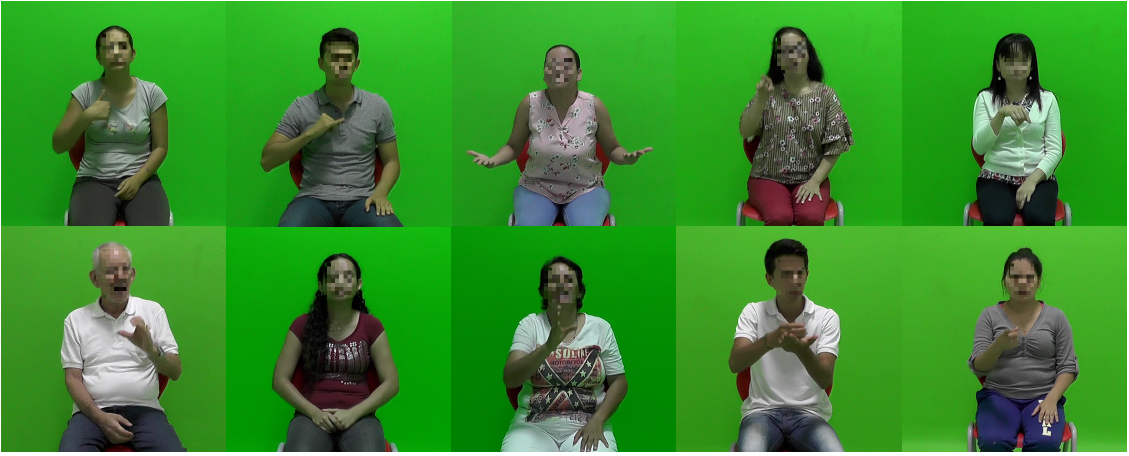}
\caption{Examples of frames from the CoL-SLTD dataset. Each video was recorded in a controlled environment, using a green screen background and ideal lighting conditions. The position of the performers is frontal to the camera, wearing different colored clothing from the background.} 
\label{fig:CoL_SLTD}
\end{figure}

The Col-SLTD dataset has videos with real and common signs of Colombian SL, with the corresponding written language translation and gloss representation. This dataset represents the first effort to quantify video signs, obtain gloss relations and record a wide variability in the eastern region of the country.  The sentences were performed by 13 interpreters between the ages of 21 and 80. A total of 24 affirmative, 4 negative and 11 interrogative sentences are part of this dataset; each sentence follows an established grammar (subject, verb, object) and was repeated 3 times by each interpreter; figure \ref{fig:CoL_SLTD} shows frames extracted from the dataset. Furthermore, the complete dataset has a vocabulary of 114 words and 90 glosses, with a total of 1020 videos. Each video has a spatial resolution of 448 $times$ 448 and a temporal resolution of 30 frames per second. In addition, CoL-SLTD has different video representations, such as: RGB, optical flow and pose estimation. 

CoL-SLTD has two data splits, which seek to evaluate the model from different ways. The first split evaluates the ability of the method to identify relevant visual features, this is achieved because the same sentences are used for both training and testing stages, varying between stages the SL sentence interpreter. The second split aims to evaluate the ability of the network to model the SL, because the same interpreters are used in both stages, varying between stages the SL sentences. Table \ref{tab:CoL_SLTD} shows the data distribution in both splits. This dataset was approved by the Ethics Committee of the Universidad Industrial de Santander in Bucaramanga, Colombia, under the number $D19-13353$.  Written informed consent was obtained from each participant.

\begin{table}[h!]
\centering

\begin{tabular}{|l|c|c|c|c|}
\hline
                      & \multicolumn{2}{|c|}{\textbf{SPLIT 1}}                 & \multicolumn{2}{|c|}{\textbf{SPLIT 2}}                 \\ \hline
                      & \multicolumn{1}{|l|}{Train} & \multicolumn{1}{|l|}{Test} & \multicolumn{1}{|l|}{Train} & \multicolumn{1}{|l|}{Test} \\ \hline
Number of videos      & 807                       & 213                      & 922                       & 98                       \\ \hline
Number of interpreters & 10                        & 3                        & 13                        & 13                       \\ \hline
Number of sentences   & 39                        & 39                       & 35                        & 4                        \\ \hline
Number of glosses      & 90                        & 90                       & 90                        & 90                       \\ \hline
Number of words    & 110                       & 110                      & 110                       & 16                      \\\hline
\end{tabular}
\caption{CoL-SLTD data distribution.}
\label{tab:CoL_SLTD}
\end{table}

\textbf{Data augmentation:} In order to obtain a more representative dataset at the training stage, a data augmentation was performed on CoL-SLTD. For this purpose, horizontal flipping strategies were employed on the video frames in order to simulate the changes of the performer's dominant hand.

\subsubsection{RWTH-PHOENIX-Weather 2014 T.}\label{subsec-RWTH}

In order to comprehensively evaluate the proposed method, one of the most widely used datasets in the state of the art and in the largest computer vision challenges was used in this work. The dataset is called RWTH-PHOENIX-Weather 2014 T and corresponds to the German SL, where 9 interpreters represent weather news through this language. RWTH-PHOENIX-Weather 2014 T has a total of 2887 words, 1066 glosses and 8257 videos, where the authors propose a subset of 7096 videos for training, 642 for testing and 519 for development. The table \ref{tab:RWTH-PHOENIX} shows the data distribution in detail.

\begin{table}[h! ]
\centering
\begin{tabular}{|l|c|c|c|}
\hline
                   & \textbf{Train} & \textbf{Dev} & \textbf{Test} \\ \hline
Number of videos   & 7096           & 519           & 642          \\ \hline
Number of words & 2887           & 951           & 1001         \\ \hline
Number of glosses    & 1066           & 393           & 411  \\ \hline   
\end{tabular}
\caption{RWTH-PHOENIX-Weather 2014 T data distribution.}
\label{tab:RWTH-PHOENIX}
\end{table}

\begin{figure}[h!].
\centering

\includegraphics[width=1\textwidth]{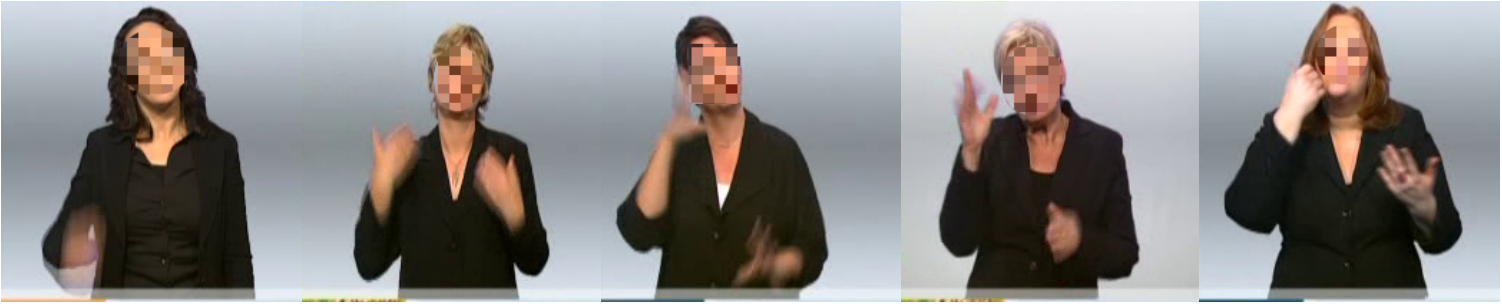}
\caption{Examples of frames from the RWTH-PHOENIX-Weather 2014 T dataset. Each video was recorded on a gray background, where each performer is wearing black clothing}. 
\label{fig:RWTH}
\end{figure}

This dataset is widely used in different SL translation strategies because it has a high information richness, making it one of the most robust for this problem. Figure \ref{fig:RWTH} shows frames extracted from the dataset.

\section{Results}

\subsection{Results in CoL-SLTD dataset}

For validation on the COL-SLTD dataset, both splits proposed by the authors were considered. Split 1 has the same sentences in training and validation, but uses different interpreters in each partition. In this way, the aim is to validate the ability of the strategies to generalize the representation of the input videos, described by different actors. For split 2, the dataset was also partitioned into training and validation, but using different sentences in both partitions. Clearly this split is more challenging, aiming for the network to generate consistency in language over unobserved representations. On the other hand, the results shown below correspond to those obtained by the beam search decoding method.

In this work, an ablation study was first carried out with the main components of the architecture. In this sense, the behavior of the proposed architecture was validated in terms of the number of attention mechanisms in each module (number of heads), the number of successive processing layers $B$ in the transformer and also the gain when processing patterns from motion vector fields (flow $(\mathbf{F})$) concerning raw sequences (RGB$(\mathbf{V})$). In this sense, the first experiment consisted of validating the proposed architecture by varying the number $C$ of heads used in the multiple attention modules. In this experiment, the transformer layers were fixed to $B=2$ for both the encoder and decoder. Table \ref{tab:compare-heads} summarizes the results obtained for the two splits of the dataset considering the different evaluation metrics. As can be seen, the number of heads directly impacts the proposed representation, obtaining, \textit{i.e.}, for split 1 a BLEU-4 with variations between $[59.26\% - 72.64\%]$ and a WER between $[39.69\% - 50.1\%]$. From these results, it can be seen that a number of $C=8$ heads achieves the best performance, while for the more challenging task of sentence generation (split 2) a larger number of heads ($C=16$) is required to get a better translation. In the split 2, there are a significantly high WER values, with no significant differences between heads, a fact associated with the complexity of sentence representation. Based on the above, the following experiments will have a fixed number $C=8$ for split 1 and $C=16$ for split 2, since these values correspond to the best translation (BLEU-4) results.

\begin{table}[h!]
\centering
\begin{tabular}{|c|c|c|c|c|c|}
\hline
\multicolumn{6}{|c|}{\textbf{SPLIT 1}}                                                                                        \\ \hline
\textbf{C} &  \textbf{WER} & \textbf{BLEU-1}         & \textbf{BLEU-2}         & \textbf{BLEU-3}         & \textbf{BLEU-4}         \\
\hline
2    &  50,1                        & 67,05          & 62,51          & 60,4          & 59,26          \\ \hline
4    & 44,55 & 69,25           & 65,69          & 64,34          & 63,83          \\ \hline
8      & 42,39 & \textbf{77,53}          & \textbf{74,42}             & \textbf{73,16}          & \textbf{72,64}          \\ \hline
16       & \textbf{39,69}            & 75,75 & 72,57 & 71,43 & 70,97 \\ 
\hline
\multicolumn{6}{|c|}{\textbf{SPLIT 2}}                                                                                        \\
\hline
2   & \textbf{81,88}                         & 33,43 & 14,97 & 10,1 & 7,45  \\ \hline
4    & 85,54                         & 23,73          & 5,98          & 2,58           & 1,55           \\ \hline
8 & 82,9                         & 30,73              & 15,62              & 10,37              & 7,52              \\ \hline
16       & 83,50 & \textbf{34,53}         & \textbf{16,21}          & \textbf{11,27}          & \textbf{8,24}           \\ \hline
\end{tabular}
\caption{Ablation study of the variable $C$ on the proposed method. The number of layers transformer corresponds to $B = 2$.}
\label{tab:compare-heads}
\end{table}

In a second experiment, the capability of the proposed method was validated based on the number of $B$ layers in the transformer. For the experiments, a symmetric transformer with an equal number of layers in both the encoder and decoder was considered. In this case, a $B_{x}^{TE}$ layer constitutes a module of multiple self-attention layers together with different processing modules, while a $B_{x}^{TD}$ layer constitutes an association of two multiple attention modules with different layers for decoding. Table \ref{tab:compare-representation} shows the results in the recognition and translation tasks in the split 1, where the performance is represented by varying the number $B$ of layers, using optical flow and raw sequences representations.

\begin{table}[h!]
\centering
\begin{tabular}{|c|c|c|c|c|c|c|}
\hline
\textbf{B}         & \textbf{Input type} & \textbf{WER}   & \multicolumn{1}{l|}{\textbf{BLEU-1}} & \multicolumn{1}{l|}{\textbf{BLEU-2}} & \multicolumn{1}{l|}{\textbf{BLEU-3}} & \multicolumn{1}{l|}{\textbf{BLEU-4}} \\ \hline
\multirow{2}{*}{1} & RGB (\textbf{V})                 & 71,81          & 42,35                                & 35,94                                & 33,45                                & 32,65                                \\ \cline{2-7} 
                   & Flow (\textbf{F})               & \textbf{48,09} & \textbf{72,44}                       & \textbf{67,7}                        & \textbf{65,43}                       & \textbf{64,34}                       \\ \hline
\multirow{2}{*}{2} & RGB (\textbf{V})                 & 63,62          & 50,82                                & 43,83                                & 41,05                                & 39,92                                \\ \cline{2-7} 
                   & Flow (\textbf{F})               & \textbf{42,39} & \textbf{77,53}                       & \textbf{74,42}                       & \textbf{73,16}                       & \textbf{72,64}                       \\ \hline
\multirow{2}{*}{3} & RGB (\textbf{V})                 & 64,97          & 51,69                                & 44,01                                & 40,97                                & 39,75                                \\ \cline{2-7} 
                   & Flow (\textbf{F})               & \textbf{35,42} & \textbf{74,65}                       & \textbf{71,58}                       & \textbf{70,48}                       & \textbf{69,95}                       \\ \hline
\multirow{2}{*}{4} & RGB (\textbf{V})                 & 61,85          & 47,75                                & 41,85                                & 39,71                                & 38,98                                \\ \cline{2-7} 
                   & Flow (\textbf{F})               & \textbf{38,67} & \textbf{75,33}                       & \textbf{72,57}                       & \textbf{71,58}                       & \textbf{71,26}                       \\ \hline
\end{tabular}
\caption{Comparison between raw video and optical flow representations in CoL-SLTD split 1. The value of $B$ corresponds to the number of transformer layers used in the experiment. For MHA modules, $8$ heads were used.}
\label{tab:compare-representation}
\end{table}

\begin{figure}[h!]
\centering

\includegraphics[width=0.9\textwidth]{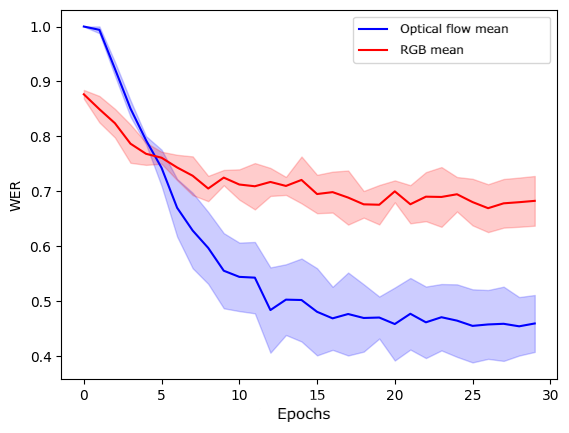}
\caption{Comparison of the WER metric for test in CoL-SLTD split 1 across the training epochs between RGB and optical flow representations.} 
\label{fig:WER_rgbflow}
\end{figure}

As expected, the optical flow sequences ($\mathbf{F}$) have a significant impact on recognition and translation tasks. In fact, in all experiments, a gain of more than $20\%$ is achieved, with respect to the raw video sequence. These differences are consistent with the hypothesis put forward in the present work, showing that a kinematic representation, which also describes the geometry of the postures, clearly impacts the representation of gestures, being less invariant to factors related to information capture. In fact, the representation gets a recognition error of about $35\%$ and a high translation quality, where the highest BLEU-4 corresponds to $72.64\%$. Also, Figure \ref{fig:WER_rgbflow} illustrates an example of the WER metric in the test stage for split 1, showing a remarkable better gloss recognition in all experiments with optical flow, where this representation drastically decreases the error in the first training epochs, guaranteeing a reliable modeling at the end of this stage. In addition, the RGB representation doesn't show considerable changes between epochs.

On the other hand, the variation of layers number in the transformer does not have an important contribution, allowing a reduced number of layers $B=2$ to obtain coherent results, with a good relation concerning the parameters needed in the computational setup. This fact may be associated with the dependence on the number of data held for training, which may not be enough for learning a high number of parameters that increase with the number of layers and may fade the representation. In this sense, starting from an optical flow representation, a compact methodology better models the SL features. Also, the best gloss recognition corresponds to $B=3$, where the optical flow presents a lower WER by $29.55\%$ compared to the RGB representation.

This same validation was performed on the split 2. Table \ref{tab:compare-representation_split2} summarizes the performance obtained by the proposed method using different input sequences and making variations in the number of layers. Similarly, most of the best results were obtained using optical flow as the input representation. However, in this case there is no significant difference between representations, where even in some cases the RGB videos obtained considerable results. In this case, the best configuration was achieved with $B=4$ layers and using optical flow sequences, achieving a $BLEU-4=14.64$ to generate coherent texts that have not been observed. It should be noted that these results are significantly lower than those obtained in the first split, due to the fact that, in this split, sentences that have not been previously observed in the training sequences are generated.

\begin{table}[h!]
\centering
\begin{tabular}{|c|c|c|c|c|c|c|}
\hline
\textbf{B}         & \textbf{Input type} & \textbf{WER}   & \multicolumn{1}{l|}{\textbf{BLEU-1}} & \multicolumn{1}{l|}{\textbf{BLEU-2}} & \multicolumn{1}{l|}{\textbf{BLEU-3}} & \multicolumn{1}{l|}{\textbf{BLEU-4}} \\ \hline
\multirow{2}{*}{1} & RGB (V)                 & 92,09          & 9,07                                 & 0                                    & 0                                    & 0                                    \\ \cline{2-7} 
                   & Flow (F)               & \textbf{75,17} & \textbf{36,73}                       & \textbf{16,02}                       & \textbf{10,18}                       & \textbf{6,96}                        \\ \hline
\multirow{2}{*}{2} & RGB (V)                 & 84,78          & 30,47                                & 11,96                                & 7,79                                 & 5,68                                 \\ \cline{2-7} 
                   & Flow (F)               & \textbf{83,5}  & \textbf{34,53}                       & \textbf{16,21}                       & \textbf{11,28}                       & \textbf{8,24}                        \\ \hline
\multirow{2}{*}{3} & RGB (V)                 & 91,83          & \textbf{28,05}                       & 8,26                                 & 4,3                                  & 2,64                                 \\ \cline{2-7} 
                   & Flow (F)               & \textbf{79,08} & 26,75                                & \textbf{8,38}                        & \textbf{4,87}                        & \textbf{2,91}                        \\ \hline
\multirow{2}{*}{4} & RGB (V)                 & 88,45          & 35,16                                & 14                                   & 9,29                                 & 6,59                                 \\ \cline{2-7} 
                   & Flow (F)               & \textbf{83,41} & \textbf{41,01}                       & \textbf{25,33}                       & \textbf{19,36}                       & \textbf{14,64}                       \\ \hline
\end{tabular}
\caption{Comparison between RGB video and optical flow representations in CoL-SLTD for split 2. The value of $B$ corresponds to the number of layers used in the experiment. For the MHA modules, $16$ heads were used.}
\label{tab:compare-representation_split2}
\end{table}

Finally, as an illustration, the table \ref{tab:translations-CoL} shows some sentences translated by the proposed method; as can be seen, the strategy is able to recognize glosses and perform quality translations. It should be noted that the translation in written language is highly based from gloss recognition, a clear example is the predicted sentence ``¿Tú cuantos años tienes?'', to which, the sequence of recognized glosses corresponds to ``TÚ QUE AÑOS?'', however, the sentence that was desired to obtain corresponds to ``¿Tú que haces?''. This shows that the proposed method takes in a relevant way the information extracted at the gloss level, using it as a guide to obtain a final translation.

\begin{table}[h!]
\centering
\begin{tabular}{|l|c|c|}
\hline
                        & \textbf{Real sentence}                                                                    & \textbf{Predicted sentence}                                                                \\
                        \hline
\textbf{Gloss}          & \begin{tabular}[c]{@{}c@{}}CARLOS VIAJAR BOGOTÁ\\  HOY\end{tabular}                    & \begin{tabular}[c]{@{}c@{}}CARLOS VIAJAR BOGOTÁ\\  HOY\end{tabular}                    \\ \hline
\textbf{Write language} & Carlos viaja a Bogotá hoy.                                                             & Carlos viaja a Bogotá hoy.                                                             \\ \hline
\textbf{Gloss}          & \begin{tabular}[c]{@{}c@{}}JUAN GUSTAR ESTO\\  Y ESO\end{tabular}                      & JUAN GUSTAR ESTO Y                                                                     \\ \hline
\textbf{Write language} & A Juan le gusta esto y eso.                                                            & A Juan le gusta esto y eso.                                                            \\ \hline
\textbf{Gloss}          & \begin{tabular}[c]{@{}c@{}}ELLA TRAER PERRO \\ COLEGIO MAÑANA\end{tabular}             & \begin{tabular}[c]{@{}c@{}}ELLA TRAER COLEGIO\\  MAÑANA\end{tabular}                   \\ \hline
\textbf{Write language} & \begin{tabular}[c]{@{}c@{}}Ella va a traer un perro\\  al colegio mañana.\end{tabular} & \begin{tabular}[c]{@{}c@{}}Ella va a traer un perro\\  al colegio mañana.\end{tabular} \\ \hline
\textbf{Gloss}          & \begin{tabular}[c]{@{}c@{}}TÚ QUE HACER?\end{tabular}             & \begin{tabular}[c]{@{}c@{}}TÚ QUE AÑOS?\end{tabular}                   \\  \hline
\textbf{Write language} & \begin{tabular}[c]{@{}c@{}}¿Tú que haces?\end{tabular} & \begin{tabular}[c]{@{}c@{}}¿Tú cuantos años\\  tienes?\end{tabular} \\ \hline
\end{tabular}
\caption[Examples of sentences translated by the proposed method in the CoL-SLTD dataset]{Examples of sentences translated by the proposed method in the CoL-SLTD dataset. In this case, the examples shown correspond to translations obtained on the split 1 test data.}
\label{tab:translations-CoL}
\end{table}

\subsubsection{Results in RWTH-PHOENIX-WEATHER 2014 T dataset}

Once the proposed strategy was evaluated on the first dataset, the hyperparameters evaluated in the testing stage were transferred to evaluate the performance on the RWTH-PHOENIX-WEATHER 2014 T dataset. For this dataset, the computational cost for training stage is very high, due to the size, so only one quantitative experiment was reported, which seeks to compare the effectiveness of the proposed method with other methodologies of the state of the art. For this purpose, the best experimental setup obtained in the previous experiments was used, which correspond to an optical flow video representation as input, a number of layers transformer $B=2$ and a number of heads $C=8$. Table \ref{tab:comparison-RWTH} shows the results obtained in the recognition and translation tasks, where the SLTT method refers to the strategy proposed in \cite{camgoz2020sign}. As can be seen, the results of the proposed method are $10.8\%$ lower in the translation task and $62.59\%$ lower in recognition. However, the proposed strategy is computationally more compact, reports a smaller number of parameters, which can be key to make a deployment in real environments. Also, the proposed strategy in this regard can better adopt and generalize the SL representation. In fact, the obtained metrics represent a remarkable BLEU-4 of $11.58\%$, which indicates that a compact strategy can potentially generalize the SL.

\begin{table}[h!]
\centering

\begin{tabular}{|c|c|c|c|c|c|}
\hline
\multicolumn{6}{|c|}{\textbf{TEST}}                                 \\ \hline
\textbf{Method}     & \textbf{WER}   & \textbf{BLEU-1} & \textbf{BLEU-2} & \textbf{BLEU-3} & \textbf{BLEU-4} \\ \hline
NSLT        & -     & 43,29  & 30,39  & 22,82  & 18,13   \\ \hline
SLTT        & \textbf{26,16} & \textbf{46,61}  & \textbf{33,73}  & \textbf{26,19}  & \textbf{21,32}  \\ \hline
Proposed approach & 89,62 & 30,55  & 20,62  & 15,09  & 11,91  \\ \hline
\multicolumn{6}{|c|}{\textbf{DEV}}                                  \\ \hline
NSLT         & -     & 42,88  & 30,30  & 23,02  & 18,40     \\ \hline
SLTT        & \textbf{24,98} & \textbf{47,26}  & \textbf{34,40}  & \textbf{27,05}  & \textbf{22,38}  \\ \hline
Proposed approach & 87,57 & 30,25  & 20,12  & 14,71  & 11,58  \\  \hline
\end{tabular}
\caption{Comparison between the proposed method and state-of-the-art methods on the RWTH-PHOENIX-WEATHER 2014 T dataset.}
\label{tab:comparison-RWTH}
\end{table}

\section{Conclusions}

In this work we presented a transformer architecture that includes multiple attention mechanisms to perform automatic translation from SL videos into textual sequences. One of the main contributions in this work was the effective inclusion of glosses as intermediate representation in video sequences and textual correspondence in the spoken language. As evidenced in the results, on two public datasets, the intermediate representation of glosses contributes to the learning of effective sign representations, as well as to obtain architectures that correctly model the SL representation. In a second contribution of the proposed work, the video sequences were processed as kinematic descriptors, first calculating the optical flow, which is projected to a volumetric convolutional architecture. From this representation it was shown to have greater strengths to the multiple variations of background and illumination, as well as allowing the inclusion of fundamental linguistic components in SL, such as kinematic primitives. With this multitask method a robust modeling of the language is achieved, performing translations at the level of written language and glosses, which are the only written approximation that correctly represents the SL.

The proposed method was thoroughly evaluated on two public datasets. As for the CoL-SLTD dataset, it demonstrated high capabilities to produce sentences with broad temporal consistency when having a training set of sentences to be predicted. Remarkable results were also achieved, surpassing the state of the art in more challenging tasks that aimed to measure the model ability to produce new sentences that are not observed in the training stage. This is evidenced by the results obtained in the BLEU and WER metrics, where the proposed method outperformed the state-of-the-art strategies in both splits of the dataset, obtaining a BLEU-4 of $72.64\%$ and a WER of $35.42\%$ for split 1 and a BLEU-4 of $14, 64\%$ and a WER of $75.17\%$ for split 2, indicating that the method can obtain efficient translations over long and short sequences, recognize reliably gestures over time, and decrease the bias when performing translations in the face of sentences not seen during training.

Likewise, the results obtained show that the use of glosses as an intermediate step can improve translation by up to $61.96\%$, indicating that the recognition of this representation has a relevant and positive impact on the quality of translations. This statement applies to both divisions of the CoL-SLTD dataset, therefore, the proposed methodology presents a better performance from different evaluative perspectives, since it can generalize gestures. It is concluded that the use of glosses as an intermediate step in SL translation contributes remarkably to language modeling, resulting in much more reliable and coherent translations. On the other hand, in the present work, it was reaffirmed that the use of motion information is a highly significant representation for language modeling, since, by employing kinematic information, superior translations of up to $32.72\%$ were obtained, this is because motion is a natural and important feature of the language.

During validation, also the proposed tool was evaluated with the RWTH-PHOENIX-WEATHER 2014 T dataset, where a competitive BLEU-4 of $11.58\%$ was obtained. The architecture achieved remarkable results and translation capabilities, being competitive for other state-of-the-art architectures, which have higher dimensionality in learning parameters. The proposed strategy in this regard is compact and robust in terms of learning capability, using a simple setting and a single hyperparameter transfer from the other dataset. Therefore, the proposed method can be easily deployed in real applications, as well as can be re-tuned with new language variabilities.

The work presented was intended to provide technological support to one of the main communication gaps of the deaf population. In addition, its implementation and validation from a Colombian dataset, aimed to approximate the modeling and coding of Colombian SL. The proposed architecture achieves a geometric representation of gestures, which is in turn enriched with kinematic information obtained from vector fields. Also, the multiple attention schemes manage to learn attention maps that represent an alternative to achieve non-linear and remote relationships in time for the encoded sequences. Despite the remarkable progress achieved with this architecture, the results show that new learning mechanisms need to be designed to deal with the wide variability of language, expressed both in textual form and gestural-visual representations. In addition, unsupervised mechanisms that can generalize representations and capture unseen sequences from minimal units of information should be explored. Finally, it is necessary to continue advancing in data collection in order to obtain a dataset with more information of Colombian SL, allowing technological mechanisms that can be deployed in real environments.

\bibliography{main}

\end{document}